\documentclass[a4paper, 10pt, conference]{ieeeconf}      % Use this line for a4 paper

\IEEEoverridecommandlockouts                              % This command is only needed if 
\usepackage{graphics} % for pdf, bitmapped graphics files
\usepackage{epsfig} % for postscript graphics files
\usepackage{mathptmx} % assumes new font selection scheme installed
\usepackage{times} % assumes new font selection scheme installed
\usepackage{amsmath} % assumes amsmath package installed
\usepackage{amssymb}  % assumes amsmath package installed
\usepackage{pgfplots}
\usepackage{import}
\usepackage{caption}
\usepackage{multirow}
\usepackage{xcolor,colortbl}
\usepackage{array}
\newcolumntype{C}[1]{>{\centering\arraybackslash}p{#1}}
\newcolumntype{G}[1]{>{\columncolor{green}\centering\arraybackslash}p{#1}}
\newcolumntype{R}[1]{>{\columncolor{red}\centering\arraybackslash}p{#1}}

\DeclareMathAlphabet{\mathcal}{OMS}{cmsy}{m}{n}

\newcommand{\vect}[1]{\pmb{#1}}  

\title{\LARGE \bf
	Deep, spatially coherent Occupancy Maps based on Radar Measurements
}

\author{Daniel Bauer$^{1}$, Lars Kuhnert$^{1}$ and Lutz Eckstein$^{2}$% <-this % stops a space
	\thanks{$^{1}$Daniel Bauer and Lars Kuhnert are  with the Ford Werke GmbH, Cologne,
		{\tt\small dbauer31@ford.de}, {\tt\small lkuhnert@ford.de}}%
	\thanks{$^{3}$Lutz Eckstein is with the Institute for Automotive Engineering, RWTH Aachen University,
		{\tt\small office@ika.rwth-aachen.de}}%
}

\begin{document}

\maketitle
\thispagestyle{empty}
\pagestyle{empty}

%%%%%%%%%%%%%%%%%%%%%%%%%%%%%%%%%%%%%%%%%%%%%%%%%%%%%%%%%%%%%%%%%%%%%%%%%%%%%%%%
\begin{abstract}
	
	One essential step to realize modern driver assistance technology is 
	the accurate knowledge about the location of static objects in the environment. 
	In this work, we use artificial neural networks to predict the occupation state of a whole scene in an end-to-end manner. 
	This stands in contrast to the traditional approach	of accumulating each detection's influence on the occupancy state and allows
	to learn spatial priors which can be used to interpolate the environment's occupancy state.
	We show that these priors make our method suitable to predict dense occupancy estimations from sparse, highly uncertain inputs, 
	as given by automotive radars, even for complex urban scenarios. Furthermore, we demonstrate that these estimations can be used
	for large-scale mapping applications.  
	
\end{abstract}

%=================================================%
%
%=================================================%
\section{Introduction}
The goal of nowadays diver assist technologies is to perform high level automation tasks like 
hazard detection, emergency breaking or path planning. To perform such tasks, a proper environment 
perception is one necessary prerequisite.
A prominent method to provide environment models for driver assistance tasks restricts itself to 
the inference of the occupation state.\\ 
The earliest approach, proposed by Elfes \cite{elfes1989using}, reduces the mapping task to a 2D problem in bird's-eye view. To ensure feasibility, they discretize the 
environment into grid cells and derive a recursive formula to update each cells occupancy state $m^{xy}$ independently.  
Following the assumptions of no prioritization of maps and the map state being a complete state \cite{thrun2005probabilistic}, the update formula for the posterior of a cell's occupancy state $m^{xy}$ given measurements $z$ can be written in logits form as follows
\begin{align}
\text{logit}( p(m^{xy}|z_{0:t}) ) = &\text{ logit}( p(m^{xy}|z_t) ) + \text{logit}( p(m^{xy}|z_{0:t-1}) )\label{eq:posterior_eq}
\end{align}
where the indices indicate whether the variables correspond to a time step or sequence.\\
Equation \eqref{eq:posterior_eq} describes an efficient, recursive update formula that uses the inverse sensor model $p(m^{xy}|z_t)$ to update the previous occupancy state estimate.
Usually, the sensor model is defined by accumulating the influences of each detection separately \cite{elfes1989using,thrun2005probabilistic,werber2015automotive}.
However, accumulation-based methods neglect the relation between detections and hence do not fully capture the spatial coherence of the scene, as described in \ref{subsec:spatial_coherence}.\\
In this paper, we propose the use of neural networks to learn a dense estimation of the occupancy state of a scene based on sparse sensor data. To do so, we first accumulate as much information about the scenes as possible by constructing occupancy maps of urban environments with LiDAR sensors. Afterwards, we use patches of the occupancy maps as labels to learn a transformation from 
\begin{center}
	%% Creator: Inkscape inkscape 0.92.3, www.inkscape.org
%% PDF/EPS/PS + LaTeX output extension by Johan Engelen, 2010
%% Accompanies image file 'architecture_overview.pdf' (pdf, eps, ps)
%%
%% To include the image in your LaTeX document, write
%%   \input{<filename>.pdf_tex}
%%  instead of
%%   \includegraphics{<filename>.pdf}
%% To scale the image, write
%%   \def\svgwidth{<desired width>}
%%   \input{<filename>.pdf_tex}
%%  instead of
%%   \includegraphics[width=<desired width>]{<filename>.pdf}
%%
%% Images with a different path to the parent latex file can
%% be accessed with the `import' package (which may need to be
%% installed) using
%%   \usepackage{import}
%% in the preamble, and then including the image with
%%   \import{<path to file>}{<filename>.pdf_tex}
%% Alternatively, one can specify
%%   \graphicspath{{<path to file>/}}
%% 
%% For more information, please see info/svg-inkscape on CTAN:
%%   http://tug.ctan.org/tex-archive/info/svg-inkscape
%%
\begingroup%
  \makeatletter%
  \providecommand\color[2][]{%
    \errmessage{(Inkscape) Color is used for the text in Inkscape, but the package 'color.sty' is not loaded}%
    \renewcommand\color[2][]{}%
  }%
  \providecommand\transparent[1]{%
    \errmessage{(Inkscape) Transparency is used (non-zero) for the text in Inkscape, but the package 'transparent.sty' is not loaded}%
    \renewcommand\transparent[1]{}%
  }%
  \providecommand\rotatebox[2]{#2}%
  \newcommand*\fsize{\dimexpr\f@size pt\relax}%
  \newcommand*\lineheight[1]{\fontsize{\fsize}{#1\fsize}\selectfont}%
  \ifx\svgwidth\undefined%
    \setlength{\unitlength}{231.30708229bp}%
    \ifx\svgscale\undefined%
      \relax%
    \else%
      \setlength{\unitlength}{\unitlength * \real{\svgscale}}%
    \fi%
  \else%
    \setlength{\unitlength}{\svgwidth}%
  \fi%
  \global\let\svgwidth\undefined%
  \global\let\svgscale\undefined%
  \makeatother%
  \begin{picture}(1,0.43014705)%
    \lineheight{1}%
    \setlength\tabcolsep{0pt}%
    \put(0,0){\includegraphics[width=\unitlength,page=1]{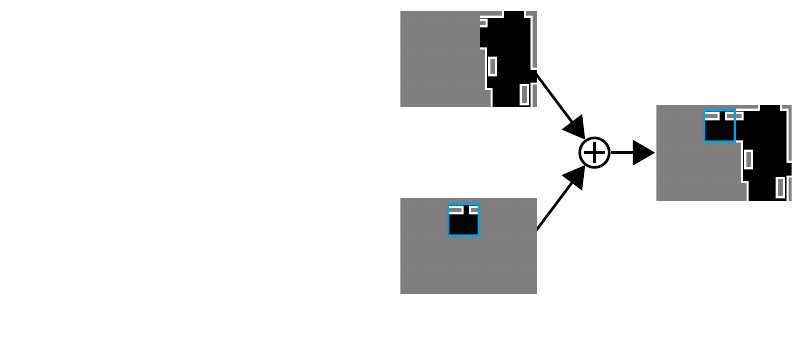}}%
    \put(0.01837627,0.18081379){\color[rgb]{0,0,0}\makebox(0,0)[lt]{\lineheight{1.25}\smash{\begin{tabular}[t]{l}\footnotesize $\vect{z}_t$\end{tabular}}}}%
    \put(0,0){\includegraphics[width=\unitlength,page=2]{architecture_overview.pdf}}%
    \put(0.25497641,0.17432881){\color[rgb]{0,0,0}\makebox(0,0)[lt]{\lineheight{1.25}\smash{\begin{tabular}[t]{l}\footnotesize $\text{logit}(\hat{\vect{y}}_t)$\end{tabular}}}}%
    \put(0,0){\includegraphics[width=\unitlength,page=3]{architecture_overview.pdf}}%
    \put(0.44884987,0.25710662){\color[rgb]{0,0,0}\makebox(0,0)[lt]{\lineheight{1.25}\smash{\begin{tabular}[t]{l}\footnotesize logit($p(m|z_{0:t-1})$)\end{tabular}}}}%
    \put(0.48775923,0.02588734){\color[rgb]{0,0,0}\makebox(0,0)[lt]{\lineheight{1.25}\smash{\begin{tabular}[t]{l}\footnotesize logit($p(m|z_t)$)\end{tabular}}}}%
    \put(0.79879562,0.14165331){\color[rgb]{0,0,0}\makebox(0,0)[lt]{\lineheight{1.25}\smash{\begin{tabular}[t]{l}\footnotesize logit($p(m|z_{0:t})$)\end{tabular}}}}%
    \put(0.08283706,0.32033485){\color[rgb]{0,0,0}\makebox(0,0)[lt]{\lineheight{1.25}\smash{\begin{tabular}[t]{l}\footnotesize Autoencoder\end{tabular}}}}%
    \put(0,0){\includegraphics[width=\unitlength,page=4]{architecture_overview.pdf}}%
    \put(0.22112016,0.11895718){\color[rgb]{0,0,0}\makebox(0,0)[lt]{\lineheight{1.25}\smash{\begin{tabular}[t]{l}\\\footnotesize map coordinates \end{tabular}}}}%
    \put(0.22112016,0.10311347){\color[rgb]{0,0,0}\makebox(0,0)[lt]{\lineheight{1.25}\smash{\begin{tabular}[t]{l}\footnotesize transform into  \end{tabular}}}}%
  \end{picture}%
\endgroup%

	
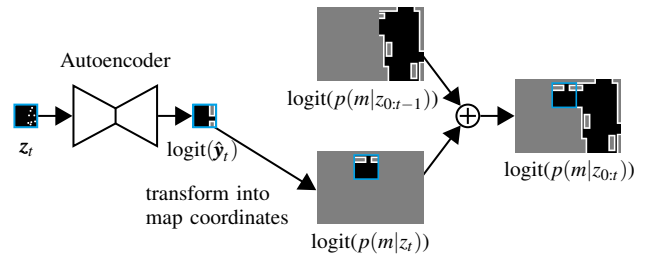
\captionof{figure}{\label{fig:architecture_overview}Architecture overview to transform sparse sensor data to dense occupancy estimates and afterwards use them to update large-scale occupancy maps. The occupancy probability is encoded as [0,1] $\rightarrow$ [black,white].}
\end{center}
sensor data to occupancy values.\\  
In our experiments, we separately train models on LiDAR and radar inputs. This enables us to compare the architectures potential to deal with both ideal and highly challenging conditions. In the end, we show that the neural network predictions can be used in a framework, illustrated in \textbf{Fig. \ref{fig:architecture_overview}}, to obtain large-scale occupancy maps which capture the underlying ground-truth.\\
To sum up, our main contributions are:
%\vspace{-2mm}
\begin{itemize}
\itemsep-0.5mm
\item[$\bullet$]
	the learning of dense, inverse sensor models applicable for sparse, highly uncertain, real-world sensors by incorporating spatial priors in an end-to-end way
\item 
	more of the sensor's information utilized by not only using the detections but also their relationship
\item[$\bullet$]
	experimental verification of the occupancy estimates by reconstructing large scale occupancy maps consistent with the ones created by traditional means based on real world data collected in an urban environment  	
\end{itemize}   
%=================================================%
%
%=================================================%
\newpage
\section{Related Work}
%================%
%
%================%
\subsection{Spatial Coherence}
\label{subsec:spatial_coherence}
One of the problems we address with our framework is the inability of accumulation-based occupancy mapping methods to capture spatial coherence. This results in 
unknown cells in areas that are highly likely to be occupied or empty vice versa, as illustrated in \textbf{Fig. \ref{fig:spatial_coherence_problem}}.
% 
%The problem is further illustrated in \textbf{Fig. \ref{fig:spatial_coherence_problem}}. Taking the area marked by the green square as an example, it is easy to see that the shape of the car and the wall behind the vehicle could be properly approximated by the manually obtained shape proposition given in blue. 
%However, because of a lack of measurements in those areas, the occupancy map results in areas of unknown state.
\begin{center}
	\includegraphics[width=2.0in,height=0.8in]{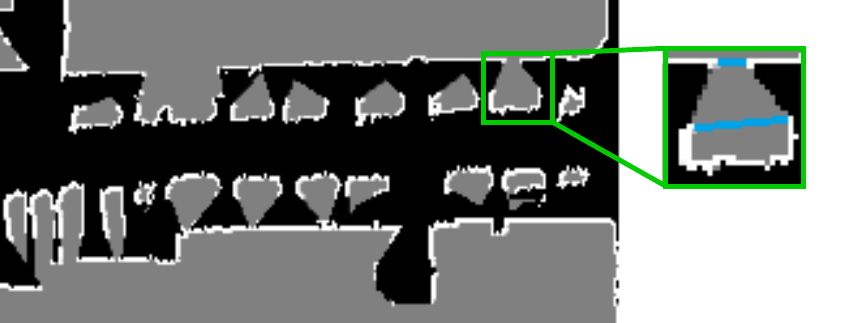}
	\captionof{figure}{\label{fig:spatial_coherence_problem}Occupancy map patch (white: occupied; gray: unknown; black: free) showing a street with parked cars.
	The anotated part shows an example, where the spatial coherence in the scene is not properly captured with a manual shape approximation (blue) for
	illustration purposes.}
\end{center}
The lack of spatial coherence in the original occupancy mapping algorithm has been addressed in several works.
One example is the work of O'Callaghan et al. \cite{o2009contextual} who applied Gaussian Processes (GPs) \cite{rasmussen2006gaussian} to the mapping problem. However, GPs are known to be computationally expensive and consume a lot of memory.\\ 
%This results in an inference method that is not restricted to the prediction of cell states but instead is capable of continuously inferring the environments state. Here, the interpolation is realized using 
%predefined kernels that weight the relation of the inference point with respect to all measurements. This procedure leads to a quadratic growth in storage and cubical growth in computational complexity with the data which restricts the use of standard GPs to small environments.\\
%
These shortcomings have been partially addressed by Kim and Kim \cite{kim2012building,kim2013continuous} in several publications. First, they proposed to cluster the data and train a separate GP for each subset which can be later used in a mixture model to perform inference. Afterwards, they proposed to use overlapping clusters with a mixture of GPs to obtain continuous inference functionality at the boarders of clusters. These solutions, however, do not get rid of the core problem and lead to many overlapping GPs in case of high resolution data or long perception ranges.\\
Another research direction was introduced by Ramos and Ott \cite{ramos2016hilbert} who propose the application of kernels to transform the input data into a Hilbert space and train a logistic regression model on the transformed data to infer the occupation state. Here, the computational expensive kernel matrix that correlates all training points with each other is approximated by a dot product of specifically designed feature functions. 
%Several extensions to this method have been proposed to for example better cope with dynamic objects \cite{senanayake2017bayesian} or incrementally fuse local Hilbert maps for large scale applications \cite{doherty2016probabilistic}.\\
%
Recently, Guizilini and Ramos proposed several enhancements of the original Hilbert mapping algorithm in \cite{guizilini2018towards} to make the method more real-time capable. These modifications include new strategies to automatically find the number of features needed for a sufficient environment description, faster methods to train the classifier and more efficient evaluation strategies.\\
Finally, Senanayake et al. \cite{senanayake2017deep} proposed the use of deep learning to obtain occupancy estimates in world coordinates based on simulated laser scanner measurements. More precisely, they build a simulation of a 2D environment and let a robot virtually drive around in this world scanning the environment using a virtual, radial, high precision sensor. These measurements are then discretized and transformed to local occupancy patches which are used as labels during the training process. The inputs of the neural network to infer those occupancy patches are the longitudinal and latitudinal positions of the cells in world coordinates. By providing longitudinal and latitudinal mesh-grids as inputs, the neural network is able to continuously infer the occupation state in the scanned environment. However, the learned transformation from position to occupancy state does not generalize. Therefore, the neural network has to be retrained from scratch for every new mapped environment. 
%================%
%
%================% 
\subsection{Radar Models}
\label{subsec:radar_models}
The second point we want to address is the capability of our model to learn complex sensor models. In this work, we concentrate on radar sensors to compute occupancy maps because they allow robust operation in various environmental conditions \cite{richards2010principles}, are capable of directly measuring distances and velocities and relatively low in cost. These abilities allow them to be used in production vehicles and make them highly relevant for today's driver assistance technologies.\\
However, proper modelling of the radar's sensor characteristics in urban scenarios is difficult for several reasons.
First, the EM wave's energy is always absorbed and scattered to a small portion by particles in the air leading to sensor noise. Furthermore, multipath reflections can
lead to falsify measurements which introduces ghost objects into the scene. These ghost objects can even have high amplitude readings caused by 
constructive interference making them hard to filter out \cite{richards2010principles}.\\ 
Moreover, radars are capable of detecting objects in 3D but lack the capability to provide height information properly. This leads to some unwanted behaviours like ground clutter which has to be accounted for.\\
Finally, the radar measurements are provided as sparse point clouds with maximal 64 points, often less, for the sensors used in this work. Thus, many radar measurements have to be accumulated or interpolated to obtain a dense prediction of the environment.\\
Wheeler et al. \cite{wheeler2017deep} have shown that it is possible to model the radar characteristics to a certain extent with deep learning approaches. More specifically, a model is trained to predict the amplitude readings of a radar given an object list and a raster grid that classifies the environment into street and grass cells. The model used was a Variational Autoencoder (VAE) conditioned on the inputs, similar as proposed in \cite{walker2016uncertain}. Additionally, to enhance the prediction quality, the VAE's loss was combined with an adversarial loss \cite{goodfellow2014generative}.          
%================%
%
%================%
\section{Deep Occupancy Maps based on Radar Measurements}
We propose a method to estimate the radar's inverse sensor model for the whole captured scene $p(m|z_t)$ in a way that incorporates prior information of the detections correlation in an end-to-end manner. We approach the problem by using radar point clouds encoded into images as inputs to an Autoencoder (AE) and trying to reconstruct the ground-truth occupancy state of the whole environment within a certain range. This ground-truth occupancy state is approximated by constructing occupancy maps and cutting out patches corresponding to the vehicle positions. By doing so, the network is able to predict occupancies in areas not in the sensor's line of sight and hence learns geometric priors.\\
Moreover, we show that the occupancy estimates can be stitched together into a global map which is consistent with maps constructed through traditional methods. Hence, we provide a framework to learn inverse sensor models capable of large-scale mapping in general urban environments.\\
Our method is based on the basic idea of \cite{wheeler2017deep} to learn the radar sensor's characteristics from data. However, while Wheeler et al. learn a forward sensor model by estimating the sensor measurements for a given environment $p(z_t|m)$, we model the inverse sensor model by estimating the environment given the sensor readings $p(m|z_t)$. These models are connected according to Bayes rule as follows
\begin{align}
	p(m|z_t) = \dfrac{p(z_t|m)p(m)}{p(z_t)}
	\label{eq:inverse_sensor_model}
\end{align}
Moreover, our method is inspired by \cite{senanayake2017deep} which however has a different focus. While they use a world referenced grid as an input to learn a continuous occupancy state function, we provide our inputs in vehicle centred coordinates. Therefore, our methods is not able to infer the occupancy at arbitrary positions but only in a fixed grid around the vehicle. However, the continuous approach has to be retrained for every new environment while our method it capable to learn to predict the occupancy state for arbitrary environments and hence can be deployed in cars more easily.
%
%To sum up, our main contributions are:
%\begin{itemize}
%\item[$\bullet$]
%	learn dense, inverse sensor models for sparse, highly uncertain, real-world sensors by incorporating spatial priors in an end-to-end way
%\item[$\bullet$]
%	experimental verification of the occupancy estimates by reconstructing large scale occupancy maps consistent with the ones created by traditional means based on real world data collected in an urban environment  	
%\end{itemize}      
%=================================================%
%
%=================================================%
\section{Experimental Setup}
%In this section, we will explain the means we used to record and process the data.
%================%
%
%================%
\subsection{Data Collection} 
The data was collected with a Lincoln MKZ equipped with four short range, automotive radars located at the corners of the car, a roof mounted Velodyne HDL-32E and the vehicle's dead-reckoning system, consisting of wheel speed and yaw rate sensors. %, as shown in \textbf{Fig. \ref{fig:lincoln_mkz_top_view}}.
%\begin{center}
%	\includegraphics[width=1.6in,height=0.8in]{img/data_collection/lincoln_mkz_top_view.pdf}
%	\captionof{figure}{\label{fig:lincoln_mkz_top_view}Test vehicle equipped with four corner radars, a roof mounted LiDAR sensor and the vehicle's odometry}
%\end{center} 
The test route, depicted in \textbf{Fig. \ref{fig:test_route}}, was planned in a way to have as few overlap as possible, while including standard, stationary, urban geometries  (e.g. parked cars, alleys, buildings, roundabouts, straight and curved road segments, etc.) in a balance way.
\begin{center}
	\includegraphics[width=2.7in,height=1.36in]{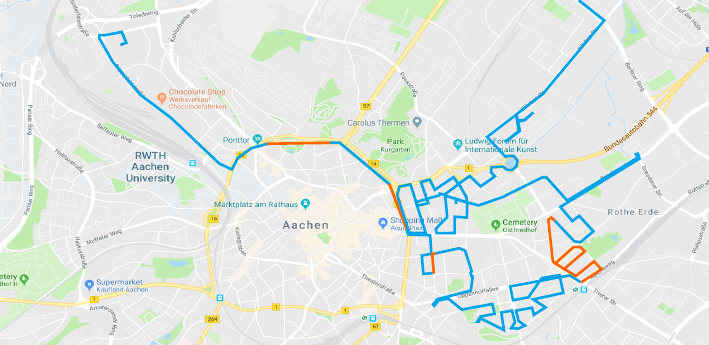}
	\captionof{figure}{\label{fig:test_route}Test route with training set marked in blue and test set marked in orange.}
\end{center}  
%================%
%
%================%
\subsection{Radar Image Patches}
The radar input images are constructed by defining an image grid for a given resolution of about $0.23m$ and perception window of $30 \times 30$ meters leading to a $128\times 128$ image. This image grid is then filled by first transforming the radar detections from polar to Cartesian coordinates and afterwards discretizing them into the image grid. In a second step, we  remove the detections corresponding to moving objects based on a threshold of the measured velocities. We explicitly do not want to incorporate the moving objects as their treatment lies beyond the scope of this work. The key characteristics of the resulting radar images are illustrated in \textbf{Fig. \ref{fig:radar_lidar_inputs_and_occupancy}}.
%================%
%
%================%
\subsection{LiDAR Image Patches}
The first step to construct the LiDAR images consists in the removal of the ground plane by applying a height threshold. Afterwards, the LiDAR's 3D point cloud is reduced to a 2D bird's eye view and only the nearest point to the vehicle for each sampled polar angle is kept. The reason for removing the other detections is that we are mainly concerned with the boundaries of the static objects in the environment. Finally, the reduced 2D point cloud is discretized into image pixels in the same way as it is done for the radar. The key characteristics of the resulting LiDAR images are illustrated in \textbf{Fig. \ref{fig:radar_lidar_inputs_and_occupancy}}.
%================%
%
%================%
\subsection{Ground-Truth-Occupancy Image Patches}
The first step to construct the ground-truth occupancy images consists of estimating the occupancy state for every reduced 2D LiDAR point cloud separately. To do so, an ideal inverse sensor model is applied for each detection where the space between the sensor and the detection is considered as free space while the detection itself indicates an occupied area.\\ 
Next, these single shot estimates are aligned using the vehicle's odometry and by fusing the overlapping parts according to Eq. \eqref{eq:posterior_eq}. Finally, the patches are cut out of the accumulated occupancy maps for each vehicle pose. The key characteristics of the resulting ground-truth occupancy images are illustrated in \textbf{Fig. \ref{fig:radar_lidar_inputs_and_occupancy}}.
The reason why we decided to use the accumulated estimates instead of the single shot estimates is to give the neural network the potential to learn shape primitives to enhance the inference capability in unobserved regions.
\begin{center}
	\def\svgwidth{2.2in}
	%% Creator: Inkscape inkscape 0.92.3, www.inkscape.org
%% PDF/EPS/PS + LaTeX output extension by Johan Engelen, 2010
%% Accompanies image file 'radar_lidar_inputs_and_occupancy.pdf' (pdf, eps, ps)
%%
%% To include the image in your LaTeX document, write
%%   \input{<filename>.pdf_tex}
%%  instead of
%%   \includegraphics{<filename>.pdf}
%% To scale the image, write
%%   \def\svgwidth{<desired width>}
%%   \input{<filename>.pdf_tex}
%%  instead of
%%   \includegraphics[width=<desired width>]{<filename>.pdf}
%%
%% Images with a different path to the parent latex file can
%% be accessed with the `import' package (which may need to be
%% installed) using
%%   \usepackage{import}
%% in the preamble, and then including the image with
%%   \import{<path to file>}{<filename>.pdf_tex}
%% Alternatively, one can specify
%%   \graphicspath{{<path to file>/}}
%% 
%% For more information, please see info/svg-inkscape on CTAN:
%%   http://tug.ctan.org/tex-archive/info/svg-inkscape
%%
\begingroup%
  \makeatletter%
  \providecommand\color[2][]{%
    \errmessage{(Inkscape) Color is used for the text in Inkscape, but the package 'color.sty' is not loaded}%
    \renewcommand\color[2][]{}%
  }%
  \providecommand\transparent[1]{%
    \errmessage{(Inkscape) Transparency is used (non-zero) for the text in Inkscape, but the package 'transparent.sty' is not loaded}%
    \renewcommand\transparent[1]{}%
  }%
  \providecommand\rotatebox[2]{#2}%
  \newcommand*\fsize{\dimexpr\f@size pt\relax}%
  \newcommand*\lineheight[1]{\fontsize{\fsize}{#1\fsize}\selectfont}%
  \ifx\svgwidth\undefined%
    \setlength{\unitlength}{224.44501742bp}%
    \ifx\svgscale\undefined%
      \relax%
    \else%
      \setlength{\unitlength}{\unitlength * \real{\svgscale}}%
    \fi%
  \else%
    \setlength{\unitlength}{\svgwidth}%
  \fi%
  \global\let\svgwidth\undefined%
  \global\let\svgscale\undefined%
  \makeatother%
  \begin{picture}(1,0.32296955)%
    \lineheight{1}%
    \setlength\tabcolsep{0pt}%
    \put(0,0){\includegraphics[width=\unitlength,page=1]{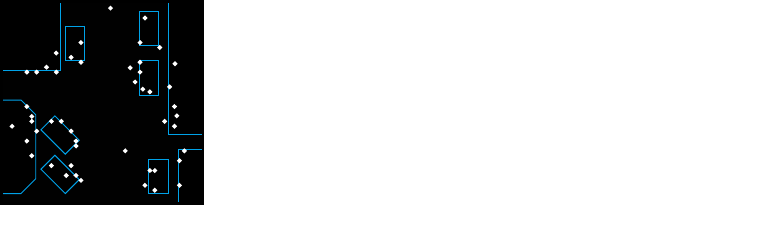}}%
    \put(0.10391102,0.01075574){\color[rgb]{0,0,0}\makebox(0,0)[lt]{\lineheight{1.25}\smash{\begin{tabular}[t]{l}a)\end{tabular}}}}%
    \put(0,0){\includegraphics[width=\unitlength,page=2]{radar_lidar_inputs_and_occupancy.pdf}}%
    \put(0.4727844,0.01075574){\color[rgb]{0,0,0}\makebox(0,0)[lt]{\lineheight{1.25}\smash{\begin{tabular}[t]{l}b)\end{tabular}}}}%
    \put(0,0){\includegraphics[width=\unitlength,page=3]{radar_lidar_inputs_and_occupancy.pdf}}%
    \put(0.8416578,0.01075574){\color[rgb]{0,0,0}\makebox(0,0)[lt]{\lineheight{1.25}\smash{\begin{tabular}[t]{l}c)\end{tabular}}}}%
  \end{picture}%
\endgroup%

	\captionof{figure}{\label{fig:radar_lidar_inputs_and_occupancy}Hand-drawn illustrations of a) radar, b) LiDAR and c) ground-truth occupancy images that show the basic characteristics of the image domains. In the radar and LiDAR images, the environment is underlayed as a reference.}
\end{center} 
%================%
%
%================%
\subsection{Data Augmentation}
To make the trained model more invariant to rotational changes we randomly rotate the input-output-pairs by random multiples of $90^\circ$ and afterwards randomly flip them along the horizontal and vertical axis.
%=================================================%
%
%=================================================%
\section{Model}
\begin{figure}
	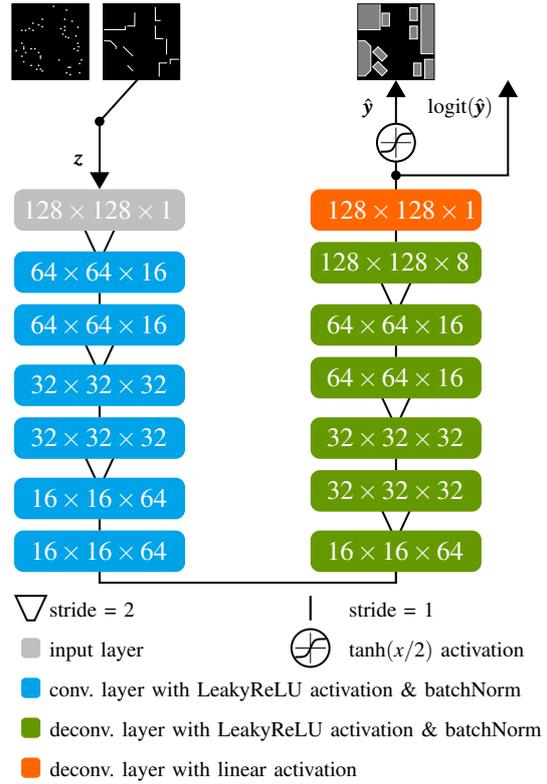
	\captionof{figure}{\label{fig:ae_architecture}Autoencoder architecture trained either with radar or LiDAR inputs.}
\end{figure}

The architecture used in this work is depicted in \textbf{Fig. \ref{fig:ae_architecture}}. As mentioned before, we use images as inputs to properly represent the spatial correlations of each detection with its surroundings. On the architectural side, convolutions are the de facto standard to learn those spatial relationships from images.\\ 
Furthermore, we decide to use an Autoencoder architecture for the following reasons. First of all, this architecture has been shown in numerous works to compress the input to lower dimensional features that capture the problem specific information. This can be used to get rid of the many unused dimensions in our inputs. Moreover, decoders like the one used in this work are the de facto standard in the transformation of latent codes into the image domain and are used for example in the DCGAN architecture \cite{radford2015unsupervised}.\\
As a final layer, a convolution layer is applied for two reasons. On the one hand, it reduces the image channels to fit the ground-truth and on the other hand it compensates the checkerboard artefacts caused by the deconvolution layers as mentioned in \cite{odena2016deconvolution}. We also experimented with the upsample-convolutional layers as proposed in \cite{odena2016deconvolution} which however only exceeded the alternative early in the training in reconstruction quality but performed slower overall.\\
To enhance the robustness and the convergence speed of the training, we linearly transform the data to be in the range of $[-1,1]$ and use LeakyRelu units as activations for the layers. Moreover, we regularized the training by using batch normalization in all layers. These methods are adapted from \cite{radford2015unsupervised,lecun2012efficient}.\\
The outputs of the last layer can be interpreted as the logits which can be used during test time in Eq. \eqref{eq:posterior_eq} to recursively compute the occupation state. However, during training, the logits have to be transformed to probabilities using the $\tanh(x/2)$ function to make them comparable with the ground-truth occupancy probabilities.
%
%================%
%
%================%
\subsection{Weighted Reconstruction Loss}
As an objective function, the mean squared error (MSE) is applied to reconstruct the ground-truth's intensity values in a continuous way. Additionally, an L2 regularizer is applied for all weights. However, this objective lacks to account for occupied areas since only about $2$\% of the pixels in the data set are defined as occupied.\\ 
Japkowicz and Stephen summarized in \cite{japkowicz2002class} several methods to deal with class imbalance. These methods can be divided into two classes.\\ 
The first class tries to sample members of the classes in a way to re-establish balance. In our case, the re-sampling could be applied by only taking a subset of the free and unknown pixels. This, however, would lead to losing spatial information and is therefore neglected.\\
The other class of methods tries to weight the loss function in a way to penalize classification errors for classes with fewer members more. For our problem, this weighted MSE with the additional L2 loss on the weights can be formulated as follows  
\begin{align}
	\mathcal{L} &= \sum_{i} \alpha_i \left \|  \hat{\vect{y}}_i-\vect{y}_i\right \|_2^2 + \lambda \sum_{j} w_j^2
\end{align}
with $\vect{y_i}$ and $\hat{\vect{y}}_i$ being ith pixel of the neural network's labels and outputs respectively, $w_j$ being the jth network weight and $\lambda$ being the regularization constant.
%======%
%
%======%
\subsubsection{Inverse Class Ratio Weighting Scheme}
In \cite{senanayake2017deep}, a weighting strategy is proposed as follows
\begin{align}
	\text{with } \alpha_i &= 
	\begin{cases}
		1-(B_f/B), \quad\text{if } \vect{y}_i = -1\\
		1-(B_u/B), \quad\text{if } \vect{y}_i = 0\\
		1-(B_o/B), \quad\text{if } \vect{y}_i = 1				
	\end{cases}
\end{align} 
with $B$ being the sum of all pixels and $B_f, B_u, B_o$ the amount of free, unknown and occupied pixels in the label image.\\  
In our case, only the occupied class has way fewer members than the free and unknown classes. This leads to the following weighting approximation
\begin{align}
	B_u &\approx B_f \approx k \cdot B_o\\
	\alpha_u &\approx \alpha_f = 1 - \dfrac{B_f}{B} = \dfrac{1+k}{1+2k} \stackrel{k \gg 1}{\approx} \dfrac{1}{2}\\
	\alpha_o &= 1 - \dfrac{B_o}{B} = \dfrac{2k}{1+2k} \stackrel{k \gg 1}{\approx} 1
\end{align}   
This means that the weighting scheme proposed in \cite{senanayake2017deep} converges for a high single-class-imbalance to a weighting scheme that halves the importance of all but the imbalanced class in the optimization.
%======%
%
%======%
\subsubsection{Independent Class MSE Weighting Scheme}
Another promising weighting scheme computes the MSE for each class separately. Afterwards, the individual losses are summed up to build the total MSE. This can be expressed in the form of a weighting scheme as follows 
\begin{align}
	\text{with } \alpha_i &= 
	\begin{cases}
		1/B_f, \quad\text{if } \vect{y}_i = -1\\
		1/B_u, \quad\text{if } \vect{y}_i = 0\\
		1/B_o, \quad\text{if } \vect{y}_i = 1				
	\end{cases} 
\end{align}
% 
%In the following section, we experiment with both the inverse class ratio and independent class MSE weighting scheme to provide a comparison of their influences for different scenarios.
%=================================================%
%
%=================================================%
\section{Experimental Results}
For our experiments, we train four different Autoencoders. Two Autoencoders based on LiDAR inputs and another two based on radar inputs with either the inverse class ratio or the independent class MSE weighting scheme as explained above.
%================%
%
%================%
\subsection{Inverse Sensor Model}
First, we want to present the learned inverse sensor models. In \textbf{Fig. \ref{fig:inverse_sensor_model_comparison_marked}}, we compare the trained LiDAR and radar models on two scenes. 
\begin{center}
	\def\svgwidth{3.2in}
	%% Creator: Inkscape inkscape 0.92.3, www.inkscape.org
%% PDF/EPS/PS + LaTeX output extension by Johan Engelen, 2010
%% Accompanies image file 'inverse_sensor_model_comparison_marked.pdf' (pdf, eps, ps)
%%
%% To include the image in your LaTeX document, write
%%   \input{<filename>.pdf_tex}
%%  instead of
%%   \includegraphics{<filename>.pdf}
%% To scale the image, write
%%   \def\svgwidth{<desired width>}
%%   \input{<filename>.pdf_tex}
%%  instead of
%%   \includegraphics[width=<desired width>]{<filename>.pdf}
%%
%% Images with a different path to the parent latex file can
%% be accessed with the `import' package (which may need to be
%% installed) using
%%   \usepackage{import}
%% in the preamble, and then including the image with
%%   \import{<path to file>}{<filename>.pdf_tex}
%% Alternatively, one can specify
%%   \graphicspath{{<path to file>/}}
%% 
%% For more information, please see info/svg-inkscape on CTAN:
%%   http://tug.ctan.org/tex-archive/info/svg-inkscape
%%
\begingroup%
  \makeatletter%
  \providecommand\color[2][]{%
    \errmessage{(Inkscape) Color is used for the text in Inkscape, but the package 'color.sty' is not loaded}%
    \renewcommand\color[2][]{}%
  }%
  \providecommand\transparent[1]{%
    \errmessage{(Inkscape) Transparency is used (non-zero) for the text in Inkscape, but the package 'transparent.sty' is not loaded}%
    \renewcommand\transparent[1]{}%
  }%
  \providecommand\rotatebox[2]{#2}%
  \newcommand*\fsize{\dimexpr\f@size pt\relax}%
  \newcommand*\lineheight[1]{\fontsize{\fsize}{#1\fsize}\selectfont}%
  \ifx\svgwidth\undefined%
    \setlength{\unitlength}{479.81694031bp}%
    \ifx\svgscale\undefined%
      \relax%
    \else%
      \setlength{\unitlength}{\unitlength * \real{\svgscale}}%
    \fi%
  \else%
    \setlength{\unitlength}{\svgwidth}%
  \fi%
  \global\let\svgwidth\undefined%
  \global\let\svgscale\undefined%
  \makeatother%
  \begin{picture}(1,0.47876893)%
    \lineheight{1}%
    \setlength\tabcolsep{0pt}%
    \put(0,0){\includegraphics[width=\unitlength,page=1]{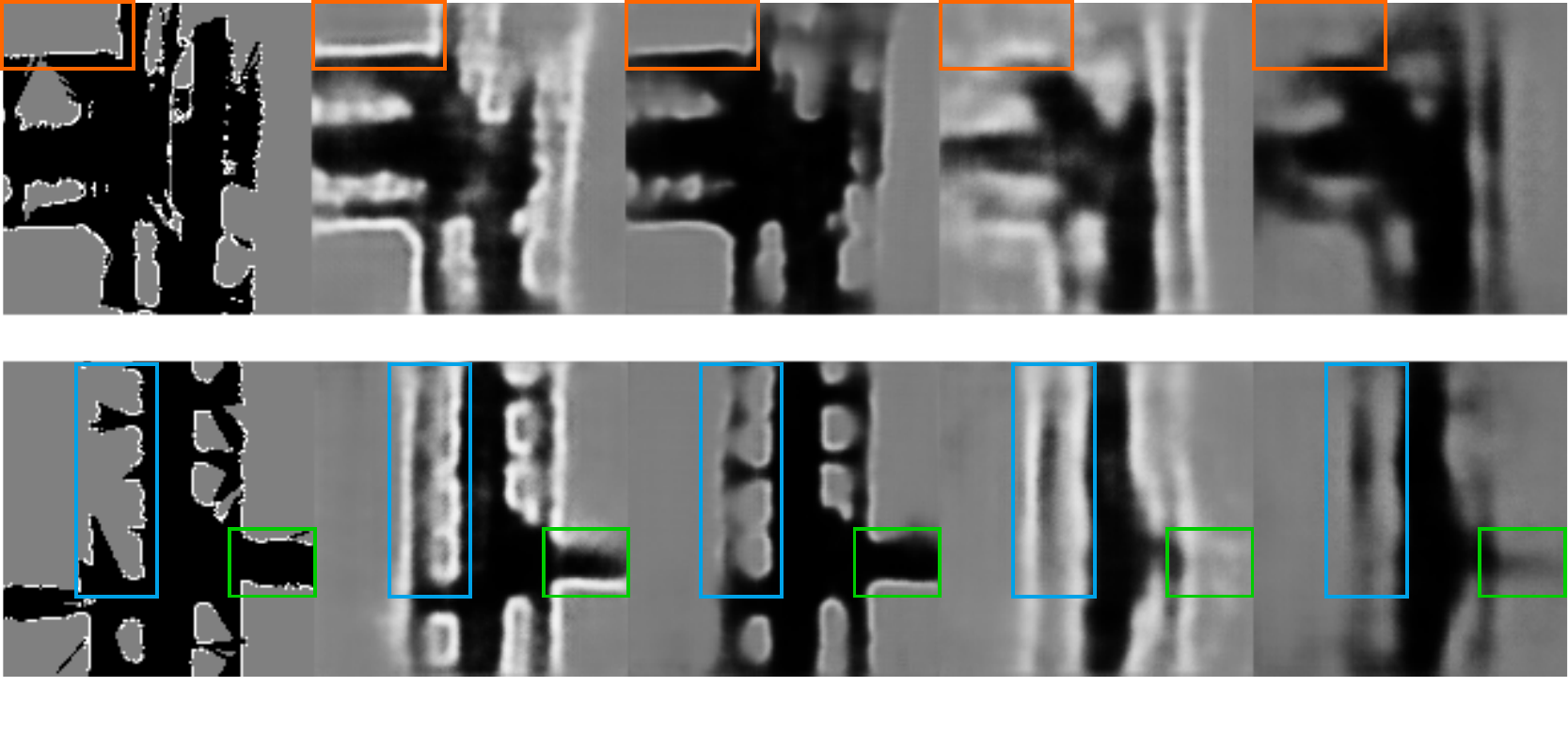}}%
    \put(0.08880025,0.00503122){\color[rgb]{0,0,0}\makebox(0,0)[lt]{\lineheight{1.25}\smash{\begin{tabular}[t]{l}a)\end{tabular}}}}%
    \put(0.28994801,0.00503122){\color[rgb]{0,0,0}\makebox(0,0)[lt]{\lineheight{1.25}\smash{\begin{tabular}[t]{l}b)\end{tabular}}}}%
    \put(0.49109577,0.00503122){\color[rgb]{0,0,0}\makebox(0,0)[lt]{\lineheight{1.25}\smash{\begin{tabular}[t]{l}c)\end{tabular}}}}%
    \put(0,0){\includegraphics[width=\unitlength,page=2]{inverse_sensor_model_comparison_marked.pdf}}%
    \put(0.69224356,0.00503122){\color[rgb]{0,0,0}\makebox(0,0)[lt]{\lineheight{1.25}\smash{\begin{tabular}[t]{l}d)\end{tabular}}}}%
    \put(0.89339129,0.00503122){\color[rgb]{0,0,0}\makebox(0,0)[lt]{\lineheight{1.25}\smash{\begin{tabular}[t]{l}e)\end{tabular}}}}%
  \end{picture}%
\endgroup%

	\captionof{figure}{\label{fig:inverse_sensor_model_comparison_marked}Comparison of the different trained inverse sensor models. The two rows show the estimation results for two different scenes. The columns show a) the ground-truth occupancy state, b) the LiDAR's inverse sensor model with the independent class MSE and c) the inverse weighting scheme respectively and in d), e) the radar pendant for the two weighting schemes.}
\end{center}
%======%
%
%======%
\subsubsection{Effects of the Weighting Schemes}
\textbf{Fig. \ref{fig:inverse_sensor_model_comparison_marked}} shows that the inverse class ratio weighting scheme is not fully able to reproduce the occupied areas indicated by white pixels. While for the LiDAR inputs the boundaries are still highlighted against the unknown and free areas, this is not the case for the radar pendant anymore.\\
In contrast, the independent class MSE weighting scheme is able to reproduce fully white boundaries even though they are less precise as compared to the LiDAR-AE with the inverse class ratio weighting. These observations are also reflected in the corresponding MSEs for occupied and free pixels, provided in \textbf{Tab. \ref{tab:class_mse}}.
\newpage
\begin{center}
\begin{tabular}{c|c|c|c} 
	& &\footnotesize Free MSE &\footnotesize Occupied MSE\\ 
\hline 
	\multirow{2}{*}{\footnotesize LiDAR} &\footnotesize inverse class ratio & 0.14 & 1.46\\
	&\footnotesize independent class MSE & 0.49 & 0.55\\ 
\hline 
	\multirow{2}{*}{\footnotesize Radar} &\footnotesize inverse class ratio & 0.19 & 1.82\\
	&\footnotesize independent class MSE & 0.70 & 0.69\\
\end{tabular} 
\captionof{table}{\label{tab:class_mse}Comparison of the mean squared reconstruction error of free and occupied cells for LiDAR and radar models trained with different weighting schemes.}
\end{center}

%======%
%
%======%
\subsubsection{Effects of Input Uncertainty}
The reason why we also trained our models on LiDAR inputs is to study in which way input uncertainty is captured in the model. By comparing the LiDAR with the radar predictions in \textbf{Fig. \ref{fig:inverse_sensor_model_comparison_marked}} one can see that the radar-AE's predictions are more "smeared" than the LiDAR's. E.g. the parked cars in a row (blue window in \textbf{Fig. \ref{fig:inverse_sensor_model_comparison_marked}}) are reconstructed as a broad line in case of the radar-AE. At the same time, the LiDAR-AE is able to reconstruct the contours pretty well. Other examples are the alley (green window) and the corner of a building (orange window) which almost can't be recognized in the radar-AE's predictions but are clearly visible for the LiDAR pendant.\\
%======%
%
%======%
\subsubsection{Learned Spatial Prior}
\textbf{Fig. \ref{fig:learned_spatial_prior}} provides a direct comparison between the scene captured by the sensor and the one learned by the model. 
\begin{center}
	\def\svgwidth{2.6in}	
	%% Creator: Inkscape inkscape 0.92.3, www.inkscape.org
%% PDF/EPS/PS + LaTeX output extension by Johan Engelen, 2010
%% Accompanies image file 'learned_spatial_prior.pdf' (pdf, eps, ps)
%%
%% To include the image in your LaTeX document, write
%%   \input{<filename>.pdf_tex}
%%  instead of
%%   \includegraphics{<filename>.pdf}
%% To scale the image, write
%%   \def\svgwidth{<desired width>}
%%   \input{<filename>.pdf_tex}
%%  instead of
%%   \includegraphics[width=<desired width>]{<filename>.pdf}
%%
%% Images with a different path to the parent latex file can
%% be accessed with the `import' package (which may need to be
%% installed) using
%%   \usepackage{import}
%% in the preamble, and then including the image with
%%   \import{<path to file>}{<filename>.pdf_tex}
%% Alternatively, one can specify
%%   \graphicspath{{<path to file>/}}
%% 
%% For more information, please see info/svg-inkscape on CTAN:
%%   http://tug.ctan.org/tex-archive/info/svg-inkscape
%%
\begingroup%
  \makeatletter%
  \providecommand\color[2][]{%
    \errmessage{(Inkscape) Color is used for the text in Inkscape, but the package 'color.sty' is not loaded}%
    \renewcommand\color[2][]{}%
  }%
  \providecommand\transparent[1]{%
    \errmessage{(Inkscape) Transparency is used (non-zero) for the text in Inkscape, but the package 'transparent.sty' is not loaded}%
    \renewcommand\transparent[1]{}%
  }%
  \providecommand\rotatebox[2]{#2}%
  \newcommand*\fsize{\dimexpr\f@size pt\relax}%
  \newcommand*\lineheight[1]{\fontsize{\fsize}{#1\fsize}\selectfont}%
  \ifx\svgwidth\undefined%
    \setlength{\unitlength}{222.72172102bp}%
    \ifx\svgscale\undefined%
      \relax%
    \else%
      \setlength{\unitlength}{\unitlength * \real{\svgscale}}%
    \fi%
  \else%
    \setlength{\unitlength}{\svgwidth}%
  \fi%
  \global\let\svgwidth\undefined%
  \global\let\svgscale\undefined%
  \makeatother%
  \begin{picture}(1,0.32443059)%
    \lineheight{1}%
    \setlength\tabcolsep{0pt}%
    \put(0,0){\includegraphics[width=\unitlength,page=1]{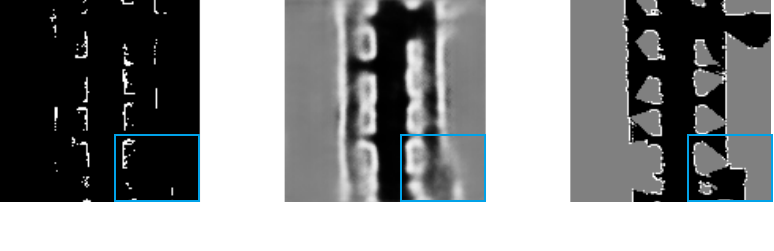}}%
    \put(0.10079318,0.01083892){\color[rgb]{0,0,0}\makebox(0,0)[lt]{\lineheight{1.25}\smash{\begin{tabular}[t]{l}a)\end{tabular}}}}%
    \put(0.47102999,0.01083892){\color[rgb]{0,0,0}\makebox(0,0)[lt]{\lineheight{1.25}\smash{\begin{tabular}[t]{l}b)\end{tabular}}}}%
    \put(0.8412667,0.01083892){\color[rgb]{0,0,0}\makebox(0,0)[lt]{\lineheight{1.25}\smash{\begin{tabular}[t]{l}c)\end{tabular}}}}%
  \end{picture}%
\endgroup%

	\captionof{figure}{\label{fig:learned_spatial_prior}Comparison between a) LiDAR input image, b) predicted occupancy state using the LiDAR-AE with the independent class MSE weighting and c) the ground-truth occupancy image}
\end{center} 
One can observe that the model is able to complete the contours of e.g. partially observed cars and walls behind them. However in areas with fewer evidence, the model predictions become less precise and tend to the unknown state (blue window).        
%================%
%
%================%
\subsection{Large Scale Mapping}
The above explained predictions of the occupancy state based on the inverse sensor model can be fused into one global map. This can be achieved by first transforming the predicted patches according to the vehicle's odometry and afterwards using Eq. \eqref{eq:posterior_eq} to fuse the overlapping parts. The result is depicted in \textbf{Fig. \ref{fig:large_scale_mapping}}.\\
Again, the sensor uncertainty is reflected in the estimations. This can for example be observed in the green window in \textbf{Fig. \ref{fig:large_scale_mapping}}, where the alley is reconstructed for the LiDAR but only partially for the radar-AE. Moreover, the parked cars can be better distinguished for the LiDAR-AE.
\begin{center}
	\def\svgwidth{3.2in}
	%% Creator: Inkscape inkscape 0.92.3, www.inkscape.org
%% PDF/EPS/PS + LaTeX output extension by Johan Engelen, 2010
%% Accompanies image file 'large_scale_mapping.pdf' (pdf, eps, ps)
%%
%% To include the image in your LaTeX document, write
%%   \input{<filename>.pdf_tex}
%%  instead of
%%   \includegraphics{<filename>.pdf}
%% To scale the image, write
%%   \def\svgwidth{<desired width>}
%%   \input{<filename>.pdf_tex}
%%  instead of
%%   \includegraphics[width=<desired width>]{<filename>.pdf}
%%
%% Images with a different path to the parent latex file can
%% be accessed with the `import' package (which may need to be
%% installed) using
%%   \usepackage{import}
%% in the preamble, and then including the image with
%%   \import{<path to file>}{<filename>.pdf_tex}
%% Alternatively, one can specify
%%   \graphicspath{{<path to file>/}}
%% 
%% For more information, please see info/svg-inkscape on CTAN:
%%   http://tug.ctan.org/tex-archive/info/svg-inkscape
%%
\begingroup%
  \makeatletter%
  \providecommand\color[2][]{%
    \errmessage{(Inkscape) Color is used for the text in Inkscape, but the package 'color.sty' is not loaded}%
    \renewcommand\color[2][]{}%
  }%
  \providecommand\transparent[1]{%
    \errmessage{(Inkscape) Transparency is used (non-zero) for the text in Inkscape, but the package 'transparent.sty' is not loaded}%
    \renewcommand\transparent[1]{}%
  }%
  \providecommand\rotatebox[2]{#2}%
  \newcommand*\fsize{\dimexpr\f@size pt\relax}%
  \newcommand*\lineheight[1]{\fontsize{\fsize}{#1\fsize}\selectfont}%
  \ifx\svgwidth\undefined%
    \setlength{\unitlength}{621.21284064bp}%
    \ifx\svgscale\undefined%
      \relax%
    \else%
      \setlength{\unitlength}{\unitlength * \real{\svgscale}}%
    \fi%
  \else%
    \setlength{\unitlength}{\svgwidth}%
  \fi%
  \global\let\svgwidth\undefined%
  \global\let\svgscale\undefined%
  \makeatother%
  \begin{picture}(1,0.59425778)%
    \lineheight{1}%
    \setlength\tabcolsep{0pt}%
    \put(0,0){\includegraphics[width=\unitlength,page=1]{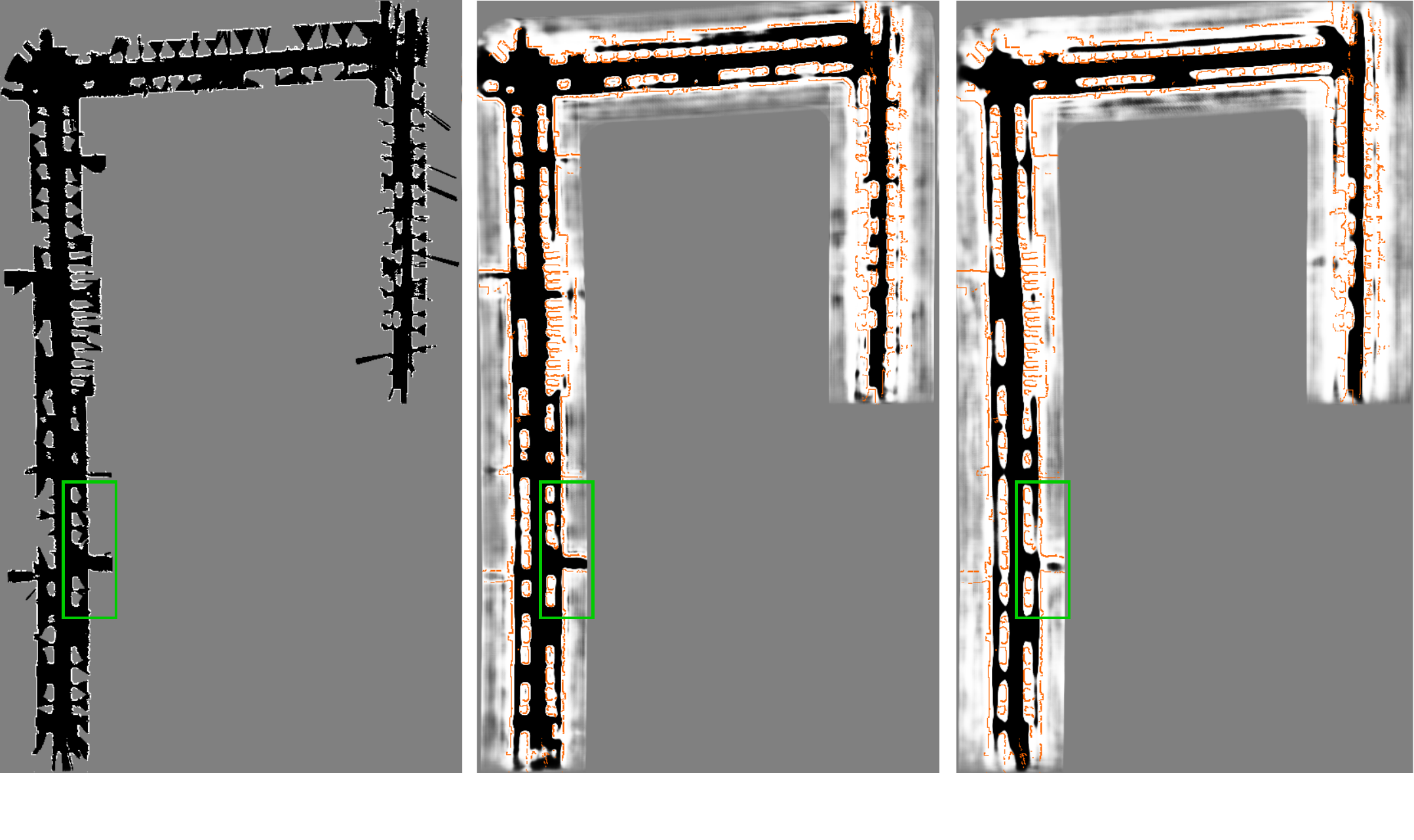}}%
    \put(0.15363241,0.00388603){\color[rgb]{0,0,0}\makebox(0,0)[lt]{\lineheight{1.25}\smash{\begin{tabular}[t]{l}a)\end{tabular}}}}%
    \put(0.49097535,0.00388603){\color[rgb]{0,0,0}\makebox(0,0)[lt]{\lineheight{1.25}\smash{\begin{tabular}[t]{l}b)\end{tabular}}}}%
    \put(0.8283183,0.00388603){\color[rgb]{0,0,0}\makebox(0,0)[lt]{\lineheight{1.25}\smash{\begin{tabular}[t]{l}c)\end{tabular}}}}%
  \end{picture}%
\endgroup%

	\captionof{figure}{\label{fig:large_scale_mapping}Comparison of a) ground-truth occupancy map, b) LiDAR and c) radar estimation. Both LiDAR and radar estimations are based on the independent class MSE weighting and are overlayed with the grounth truth occupancy estimations in orange.}
\end{center}
%=================================================%
%
%=================================================%
\section{Conclusion}
In this work, we have demonstrated the capability of Autoencoders to learn inverse sensor models to capture the boundaries of static objects in an environment. The experiments have shown that the architecture can handle highly uncertain, sparse input data as provided by automotive radar sensors and is still able to predict the environment in a way that captures the underlying geometries spatially coherent. Moreover, we have demonstrated that the model can be used for large-scale mapping tasks in complex urban environments.
%=================================================%
%
%=================================================%
\section{Acknowledgements}
We like to thank Praveen Narayanan and Punarjay Chakravarty for the insightful discussions and guiding remarks which let to great improvements of this work.  
%=================================================%
%
%=================================================%
\bibliographystyle{plain}
\bibliography{bib}
\end{document}